\documentclass{article}
\usepackage[T1]{fontenc}
\usepackage[utf8]{inputenc}
\usepackage{url}
\usepackage{amsmath}
\usepackage{graphicx}
\usepackage[unicode=true,
 bookmarks=false,
 breaklinks=false,pdfborder={0 0 1},backref=section,colorlinks=false]
 {hyperref}

\makeatletter

\providecommand{\tabularnewline}{\\}

\usepackage[final, nonatbib]{nips_2016}

\usepackage{times}
\usepackage{subfigure}
\usepackage{amsfonts}
\usepackage{cite}
\usepackage{color}
\usepackage{url}
\usepackage{wrapfig}

\usepackage{algorithm}
\usepackage{algorithmic}

\usepackage{url}
\usepackage{booktabs}
\usepackage{amsfonts}
\usepackage{nicefrac}
\usepackage{microtype}
\usepackage{wrapfig}
\usepackage{lipsum}
\usepackage{booktabs}

\makeatother

\begin{document}

\title{Using Social Dynamics to Make Individual Predictions:\\
Variational Inference with a Stochastic Kinetic Model}

\author{Zhen Xu, Wen Dong, and Sargur Srihari\\
Department of Computer Science and Engineering\\
University at Buffalo\\
\texttt{\{zxu8,wendong,srihari\}@buffalo.edu}}
\maketitle
\begin{abstract}
Social dynamics is concerned primarily with interactions among individuals
and the resulting group behaviors, modeling the temporal evolution
of social systems via the interactions of individuals within these
systems. In particular, the availability of large-scale data from
social networks and sensor networks offers an unprecedented opportunity
to predict state-changing events at the individual level. Examples
of such events include disease transmission, opinion transition in
elections, and rumor propagation. Unlike previous research focusing
on the collective effects of social systems, this study makes efficient
inferences at the individual level. In order to cope with dynamic
interactions among a large number of individuals, we introduce the
stochastic kinetic model to capture adaptive transition probabilities
and propose an efficient variational inference algorithm the complexity
of which grows \emph{linearly} — rather than exponentially— with the
number of individuals. To validate this method, we have performed
epidemic-dynamics experiments on wireless sensor network data collected
from more than ten thousand people over three years. The proposed
algorithm was used to track disease transmission and predict the probability
of infection for each individual. Our results demonstrate that this
method is more efficient than sampling while nonetheless achieving
high accuracy. 
\end{abstract}

\section{Introduction}

\label{intro}The field of social dynamics is concerned primarily
with interactions among individuals and the resulting group behaviors.
Research in social dynamics models the temporal evolution of social
systems via the interactions of the individuals within these systems
\cite{DuYo04}. For example, opinion dynamics can model the opinion
state transitions of an entire population in an election scenario
\cite{CaFo09}, and epidemic dynamics can predict disease outbreaks
ahead of time \cite{EuGu04}. While traditional social-dynamics models
focus primarily on the macroscopic effects of social systems, often
we instead wish to know the answers to more specific questions. Given
the movement and behavior history of a subject with Ebola, can we
tell how many people should be tested or quarantined? City-size quarantine
is not necessary, but family-size quarantine is insufficient. We aim
to model a method to evaluate the paths of illness transmission and
the risks of infection for \emph{individuals}, so that limited medical
resources can be most efficiently distributed.

The rapid growth of both social networks and sensor networks offers
an unprecedented opportunity to collect abundant data at the individual
level. From these data we can extract temporal interactions among
individuals, such as meeting or taking the same class. To take advantage
of this opportunity, we model social dynamics from an individual perspective.
Although such an approach has considerable potential, in practice
it is difficult to model the dynamic interactions and handle the costly
computations when a large number of individuals are involved. In this
paper, we introduce an event-based model into social systems to characterize
their temporal evolutions and make tractable inferences on the individual
level.

Our research on the temporal evolutions of social systems is related
to dynamic Bayesian networks and continuous time Bayesian networks
\cite{RoJo10,HeGh13,NoSh02}. Traditionally, a coupled hidden Markov
model is used to capture the interactions of components in a system
\cite{BrOl97}, but this model does not consider dynamic interactions.
However, a stochastic kinetic model is capable of successfully describing
the interactions of molecules (such as collisions) in chemical reactions
\cite{WiDa11,GoWi11}, and is widely used in many fields such as chemistry
and cell biology \cite{ArRo98,GiDa07}. We introduce this model into
social dynamics and use it to focus on individual behaviors. 

A challenge in capturing the interactions of individuals is that in
social dynamics the state space grows exponentially with the number
of individuals, which makes exact inference intractable. To resolve
this we must apply approximate inference methods. One class of these
involves sampling-based methods. Rao and Teh introduce a Gibbs sampler
based on local updates \cite{NRaTe12}, while Murphy and Russell introduce
Rao-Blackwellized particle filtering for dynamic Bayesian networks
\cite{MuRu01}. However, sampling-based methods sometimes mix slowly
and require a large number of samples/particles. To demonstrate this
issue, we offer empirical comparisons with two major sampling methods
in Section \ref{sec:experiments}. An alternative class of approximations
is based on variational inference. Opper and Sanguinetti apply the
variational mean field approach to factor a Markov jump process \cite{OpSa08},
and Cohn and El-Hay further improve its efficiency by exploiting the
structure of the target network \cite{CoEl10}. A problem is that
in an event-based model such as a stochastic kinetic model (SKM),
the variational mean field is not applicable when a single event changes
the states of two individuals simultaneously. Here, we use a general
expectation propagation principle \cite{HeZo02} to design our algorithm. 

This paper makes three contributions: First, we introduce the discrete
event model into social dynamics and make tractable inferences on
both individual behaviors and collective effects. To this end, we
apply the stochastic kinetic model to define adaptive transition probabilities
that characterize the dynamic interaction patterns in social systems.
Second, we design an efficient variational inference algorithm whose
computation complexity grows linearly with the number of individuals.
As a result, it scales very well in large social systems. Third, we
conduct experiments on epidemic dynamics to demonstrate that our algorithm
can track the transmission of epidemics and predict the probability
of infection for each individual. Further, we demonstrate that the
proposed method is more efficient than sampling while nonetheless
achieving high accuracy.

The remainder of this paper is organized as follows. In Section 2,
we briefly review the coupled hidden Markov model and the stochastic
kinetic model. In Section 3, we propose applying a variational algorithm
with the stochastic kinetic model to make tractable inferences in
social dynamics. In Section 4, we detail empirical results from applying
the proposed algorithm to our epidemic data along with the proximity
data collected from sensor networks. Section 5 concludes. 

\section{Background}

\subsection{Coupled Hidden Markov Model}

\label{sec:CHMM}
\begin{figure}
\label{CHMMvsSKM}\center \hfil\subfigure[]{\includegraphics[width=0.36\columnwidth]{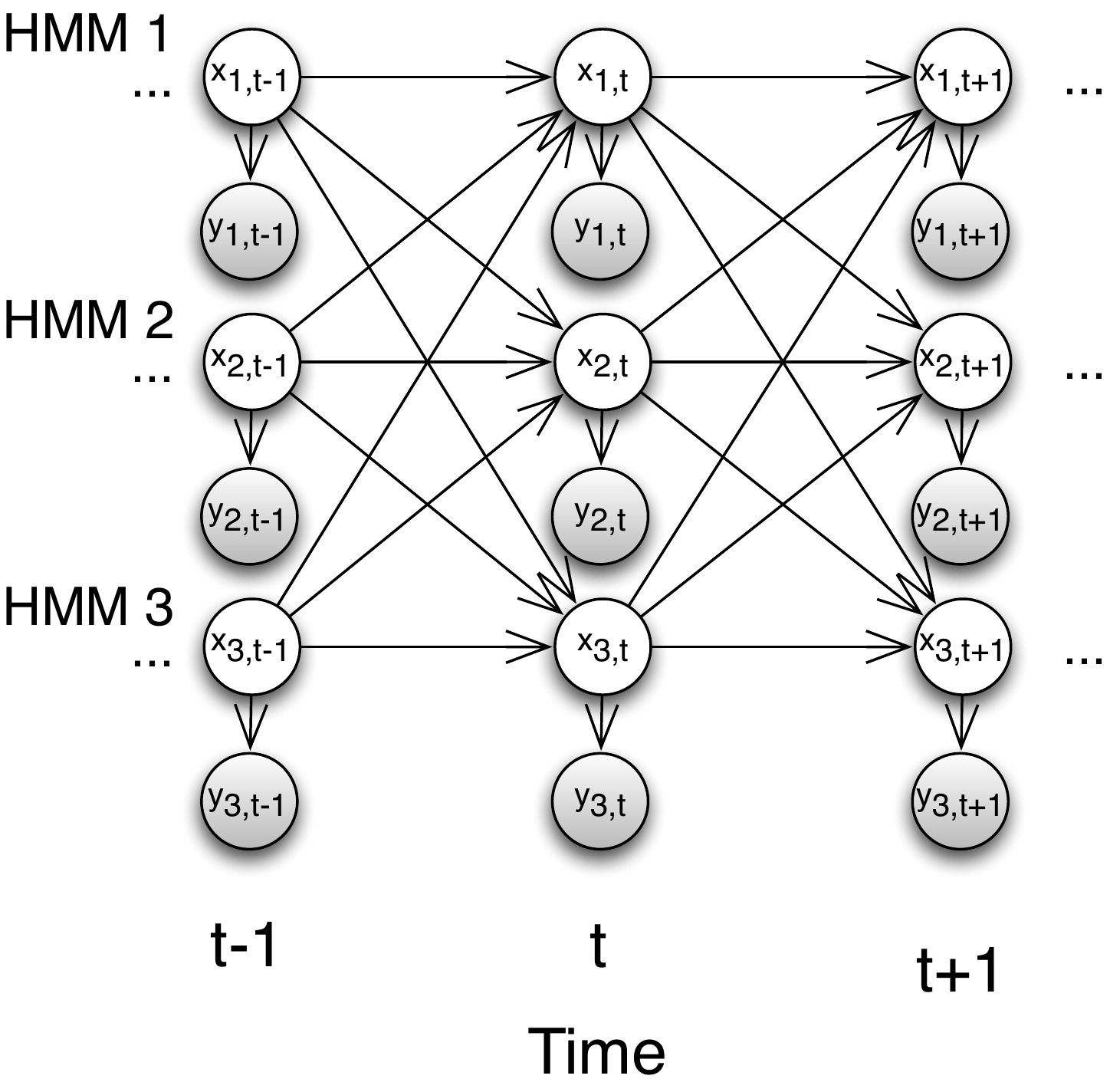}}
\hfil\subfigure[]{\includegraphics[width=0.36\columnwidth]{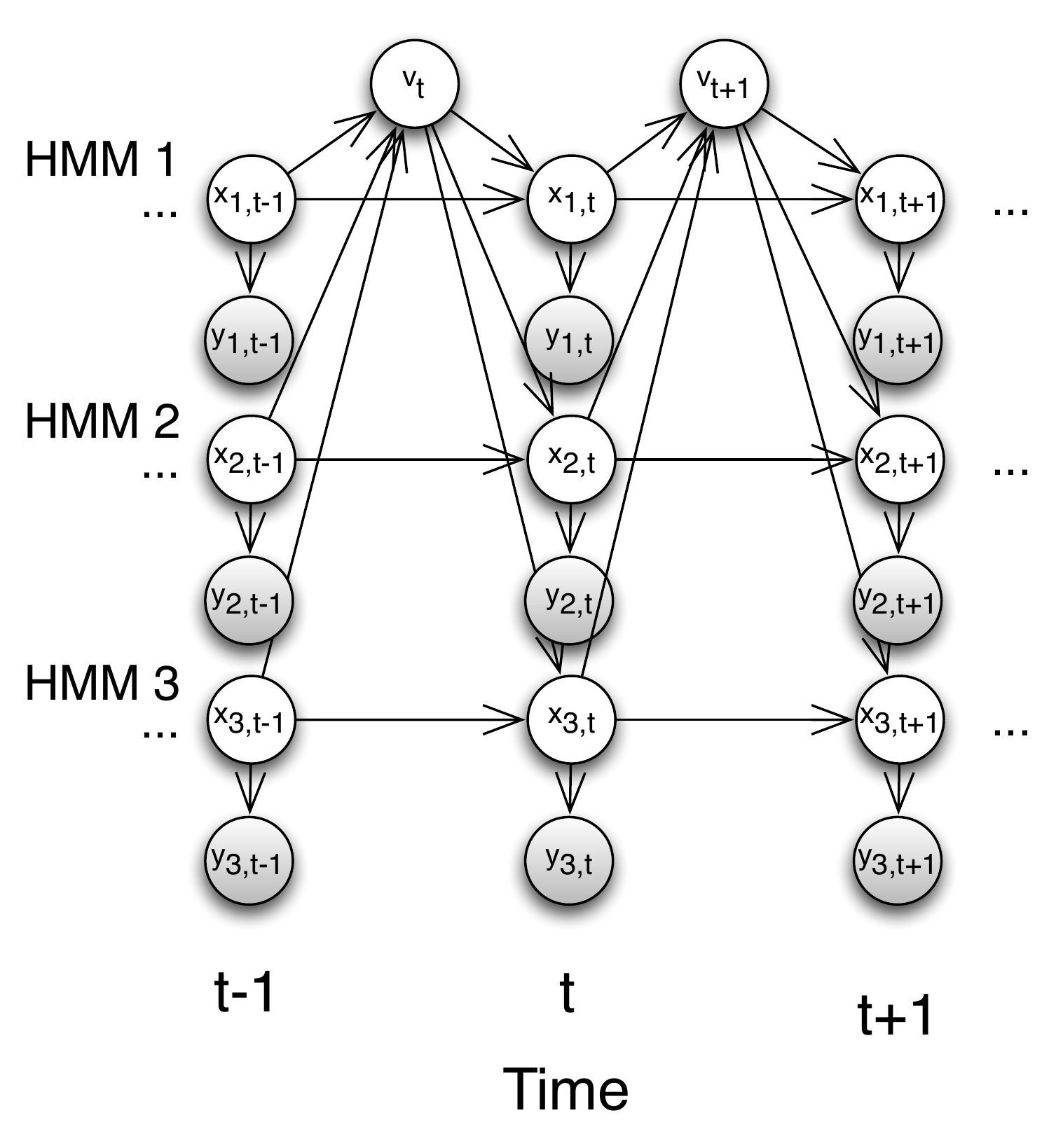}}\hfil
\protect\caption{Illustration of (a) Coupled Hidden Markov Model, (b) Stochastic Kinetic
Model.}
\end{figure}

A coupled hidden Markov model (CHMM) captures the
dynamics of a discrete time Markov process that joins a number of
distinct hidden Markov models (HMMs), as shown in Figure \ref{CHMMvsSKM}(a).
$\mathbf{x}_{t}=(x_{t}^{(1)},\dots,x_{t}^{(M)})$ defines the hidden
states of all HMMs at time $t$, and $x_{t}^{(m)}$ is the hidden
state of HMM $m$ at time $t$. $\mathbf{y}_{t}=(y_{t}^{(1)},\dots,y_{t}^{(M)})$
are observations of all HMMs at time $t$, and $y_{t}^{(m)}$ is the
observation of HMM $m$ at time $t$. $P(\mathbf{x}_{t}|\mathbf{x}_{t-1})$
are transition probabilities, and $P(\mathbf{y}_{t}|\mathbf{x}_{t})$
are emission probabilities for CHMM. Given hidden states, all observations
are independent. As such, $P(\mathbf{y}_{t}|\mathbf{x}_{t})=\prod_{m}P(y_{t}^{(m)}|x_{t}^{(m)})$,
where $P(y_{t}^{(m)}|x_{t}^{(m)})$ is the emission probability for
HMM $m$ at time $t$. The joint probability of CHMM can be defined
as follows: 
\begin{equation}
P\left(\mathbf{x}_{1,\dots,T},\mathbf{y}_{1,\dots,T}\right)=\prod_{t=1}^{T}P(\mathbf{x}_{t}|\mathbf{x}_{t-1})P(\mathbf{y}_{t}|\mathbf{x}_{t}).\label{eq:likelihood}
\end{equation}
For a CHMM that contains $M$ HMMs in a binary state, the state space
is $2^{M}$, and the state transition kernel is a $2^{M}\times2^{M}$
matrix. In order to make exact inferences, the classic forward-backward
algorithm sweeps a forward/filtering pass to compute the forward statistics
$\alpha_{t}(\mathbf{x}_{t})=P(\mathbf{x}_{t}|\mathbf{y}_{1,\dots,t})$
and a backward/smoothing pass to estimate the backward statistics
$\beta_{t}(\mathbf{x}_{t})=\frac{P(\mathbf{y}_{t+1,\dots,T}|\mathbf{x}_{t})}{P(\mathbf{y}_{t+1,\dots,T}|\mathbf{y}_{1,\dots,t})}$.
Then it can estimate the one-slice statistics $\gamma_{t}(\mathbf{x}_{t})=P(\mathbf{x}_{t}|\mathbf{y}_{1,\dots,T})=\alpha_{t}(\mathbf{x}_{t})\beta_{t}(\mathbf{x}_{t})$
and two-slice statistics $\xi_{t}(\mathbf{x}_{t-1},\mathbf{x}_{t})=P(\mathbf{x}_{t-1},\mathbf{x}_{t}|\mathbf{y}_{1,\dots,T})=\frac{\alpha_{t-1}(\mathbf{x}_{t-1})P(\mathbf{x}_{t}|\mathbf{x}_{t-1})P(\mathbf{y}_{t}|\mathbf{x}_{t})\beta_{t}(\mathbf{x}_{t})}{P(\mathbf{y}_{t}|\mathbf{y}_{1,\dots,t-1})}$.
Its complexity grows exponentially with the number of HMM chains.
In order to make tractable inferences, certain factorizations and
approximations must be applied. In the next section, we introduce
a stochastic kinetic model to lower the dimensionality of transition
probabilities.

\subsection{The Stochastic Kinetic Model}

A stochastic kinetic model describes the temporal evolution of a chemical
system with $M$ species $\mathcal{X}=\{X_{1},X_{2},\cdots,X_{M}\}$
driven by $V$ events (or chemical reactions) parameterized by rate
constants $\mathbf{c}=(c_{1},\dots,c_{V})$. An event (chemical reaction)
$k$ has a general form as follows: 
\begin{eqnarray*}
r_{1}X_{1}+\cdots+r_{M}X_{M}\overset{c_{k}}{\longrightarrow}p_{1}X_{1}+\cdots+p_{M}X_{M}.
\end{eqnarray*}
The species on the left are called \emph{reactants}, and $r_{m}$
is the number of $m$th reactant molecules consumed during the reaction.
The species on the right are called \emph{products}, and $p_{m}$
is the number of $m$th product molecules produced in the reaction.
Species involved in the reaction ($r_{m}>0$) without consumption
or production ($r_{m}=p_{m}$) are called \emph{catalysts}. At any
specific time $t$, the populations of the species is $\mathbf{x_{t}}=(x_{t}^{(1)},\dots,x_{t}^{(M)})$.
An event $k$ happens with rate $h_{k}(\mathbf{x_{t}},c_{k})$, determined
by the rate constant and the current population state \cite{WiDa11}:
\begin{align}
h_{k}(\mathbf{x_{t}},c_{k})= & c_{k}g_{k}(\mathbf{x_{t}})=c_{k}\prod_{m=1}^{M}g_{k}^{(m)}(x_{t}^{(m)}).\label{eq:rate}
\end{align}
The form of $g_{k}(\mathbf{x_{t}})$ depends on the reaction. In our
case, we adopt the product form $\prod_{m=1}^{M}g_{k}^{(m)}(x_{t}^{(m)})$,
which represents the total number of ways that reactant molecules
can be selected to trigger event $k$ \cite{WiDa11}. Event $k$ changes
the populations by $\mathbf{\Delta_{k}}=\mathbf{x}_{t}-\mathbf{x}_{t-1}$.
The probability that event $k$ will occur during time interval $(t,t+dt]$
is $h_{k}(\mathbf{x_{t}},c_{k})dt$. We assume at each discrete time
step that no more than one event will occur. This assumption follows
the linearization principle in the literature \cite{NoSh02}, and
is valid when the discrete time step is small. We treat each discrete
time step as a unit of time, so that $h_{k}(\mathbf{x_{t}},c_{k})$
represents the probability of an event.

In epidemic modeling, for example, an infection event $v_{i}$ has
the form $S+I\overset{c_{i}}{\longrightarrow}2I$, such that a susceptible
individual ($S$) is infected by an infectious individual ($I$) with
rate constant $c_{i}$. If there is only one susceptible individual
(type $m=1$) and one infectious individual (type $m=2$) involved
in this event, $h_{i}(\mathbf{x_{t}},c_{i})=c_{i}$, $\mathbf{\Delta_{i}}=[-1~~1]^{T}$
and $P(\mathbf{x}_{t}-\mathbf{x}_{t-1}=\mathbf{\Delta_{i}})=P(\mathbf{x}_{t}|\mathbf{x}_{t-1},v_{i})=c_{i}$.

In a traditional hidden Markov model, the transition kernel is typically
fixed. In comparison, SKM is better at capturing dynamic interactions
in terms of the events with rates dependent on reactant populations,
as shown in Eq.(\ref{eq:rate}).

\section{Variational Inference with the Stochastic Kinetic Model}

\vspace{-5pt}
In this section, we define the likelihood of the entire sequence of
hidden states and observations for an event-based model, and derive
a variational inference algorithm and parameter-learning algorithm.\vspace{-5pt}

\subsection{Likelihood for Event-based Model}

\vspace{-5pt}
In social dynamics, we use a discrete time Markov model to describe
the temporal evolutions of a set of individuals $x^{(1)},\dots,x^{(M)}$
according to a set of $V$ events. To cope with dynamic interactions,
we introduce the SKM and express the state transition probabilities
in terms of event probabilities, as shown in Figure \ref{CHMMvsSKM}(b).
We assume at each discrete time step that no more than one event will
occur. Let $v_{1},\dots,v_{T}$ be a sequence of events, $\mathbf{x_{1}},\dots,\mathbf{x_{T}}$
a sequence of hidden states, and $\mathbf{y_{1}},\dots,\mathbf{y_{T}}$
a set of observations. Similar to Eq.(\ref{eq:likelihood}), the likelihood
of the entire sequence is as follows: 
\begin{align}
 & P\left(\mathbf{x}_{1,\dots,T},\mathbf{y}_{1,\dots,T},v_{1,\dots,T}\right)=\prod_{t=1}^{T}P(\mathbf{x}_{t},v_{t}|\mathbf{x}_{t-1})P(\mathbf{y}_{t}|\mathbf{x}_{t}),\mbox{ where }\label{eq:DTSKM}\\
 & P(\mathbf{x}_{t},v_{t}|\mathbf{x}_{t-1})=\begin{cases}
c_{k}\cdot g_{k}\left(\mathbf{x}_{t-1}\right)\cdot\delta(\mathbf{x}_{t}-\mathbf{x}_{t-1}\equiv\mathbf{\Delta_{k}}) & \mbox{if }v_{t}=k\\
(1-\sum_{k}c_{k}g_{k}\left(\mathbf{x}_{t-1}\right))\cdot\delta(\mathbf{x}_{t}-\mathbf{x}_{t-1}\equiv\mathbf{0}) & \mbox{if }v_{t}=\emptyset
\end{cases}.
\nonumber 
\end{align}
$P(\mathbf{x}_{t},v_{t}|\mathbf{x}_{t-1})$ is the event-based transition
kernel. $\delta(\mathbf{x}_{t}-\mathbf{x}_{t-1}\equiv\mathbf{\Delta_{k}})$
is 1 if the previous state is $\mathbf{x}_{t-1}$ and the current
state is $\mathbf{x}_{t}=\mathbf{x}_{t-1}+\mathbf{\Delta_{k}}$, and
0 otherwise. $\mathbf{\Delta_{k}}$ is the effect of event $v_{k}$.
$\emptyset$ represents an auxiliary event, meaning that there is
no event. Substituting the product form of $g_{k}$, the transition
kernel can be written as follows: 
\begin{align}
 & P(\mathbf{x}_{t},v_{t}=k|\mathbf{x}_{t-1})=c_{k}\prod_{m}g_{k}^{(m)}(x_{t-1}^{(m)})\cdot\prod_{m}\delta(x_{t}^{(m)}-x_{t-1}^{(m)}\equiv\Delta_{k}^{(m)}),\label{eq:SKMEventRate}\\
 & P(\mathbf{x}_{t},v_{t}=\emptyset|\mathbf{x}_{t-1})=(1-\sum_{k}c_{k}\prod_{m}g_{k}^{(m)}(x_{t-1}^{(m)}))\cdot\prod_{m}\delta(x_{t}^{(m)}-x_{t-1}^{(m)}\equiv0),\label{eq:SKMNERate}
\end{align}
where $\delta(x_{t}^{(m)}-x_{t-1}^{(m)}\equiv\Delta_{k}^{(m)})$ is
1 if the previous state of an individual $m$ is $x_{t-1}^{(m)}$
and the current state is $x_{t}^{(m)}=x_{t-1}^{(m)}+\Delta_{k}^{(m)}$,
and 0 otherwise. \vspace{-5pt}

\subsection{Variational Inference for Stochastic Kinetic Model}

\label{sec:variational_inference} \vspace{-5pt}
As noted in Section \ref{sec:CHMM}, exact inference in social dynamics
is intractable due to the formidable state space. However, we can
approximate the posterior distribution $P(\mathbf{x}_{1,...,T},v_{1,...,T}|\mathbf{y}_{1,...,T})$
using an approximate distribution within the exponential family. The
inference algorithm minimizes the KL divergence between these two
distributions, which can be formulated as an optimization problem
\cite{HeZo02}: 
\begin{align}
 & \mbox{Minimize:}\sum_{t,\mathbf{x}_{t-1},\mathbf{x}_{t},v_{t}}\hat{\xi}(\mathbf{x}_{t-1},\mathbf{x}_{t},v_{t})\cdot\log\frac{\hat{\xi}(\mathbf{x}_{t-1},\mathbf{x}_{t},v_{t})}{P(\mathbf{x}_{t},v_{t}|\mathbf{x}_{t-1})P(\mathbf{y}_{t}|\mathbf{x}_{t})}\label{eq:BetheEnergy}\\
 & \hspace*{18em}-\sum_{t,\mathbf{x}_{t}}\prod_{m}\hat{\gamma}_{t}^{(m)}(x_{t}^{(m)})\log\prod_{m}\hat{\gamma}_{t}^{(m)}(x_{t}^{(m)})\nonumber \\
 & \mbox{Subject to: }\sum_{v_{t},\mathbf{x}_{t-1},\{\mathbf{x}_{t}\backslash x_{t}^{(m)}\}\hidewidth}\hat{\xi}(\mathbf{x}_{t-1},\mathbf{x}_{t},v_{t})=\hat{\gamma}_{t}^{(m)}(x_{t}^{(m)})\mbox{, for all }t,m,x_{t}^{(m)},\nonumber \\
 & \hphantom{\mbox{Subject to: }}\sum_{v_{t},\{\mathbf{x}_{t-1}\backslash x_{t-1}^{(m)}\},\mathbf{x}_{t}\hidewidth}\hat{\xi}(\mathbf{x}_{t-1},\mathbf{x}_{t},v_{t})=\hat{\gamma}_{t-1}^{(m)}(x_{t-1}^{(m)})\mbox{, for all }t,m,x_{t-1}^{(m)},~\nonumber \\
 & \hphantom{\mbox{Subject to: }}\sum_{x_{t}^{(m)}\hidewidth}\hat{\gamma}_{t}^{(m)}(x_{t}^{(m)})=1\mbox{, for all }t,m.\nonumber 
\end{align}
The objective function is the Bethe free energy, composed of average
energy and Bethe entropy approximation \cite{YeFr03}. $\hat{\xi}(\mathbf{x}_{t-1},\mathbf{x}_{t},v_{t})$
is the approximate two-slice statistics and $\hat{\gamma}^{(m)}(x_{t}^{(m)})$
is the approximate one-slice statistics for each individual $m$.
They form the approximate distribution over which to minimize the
Bethe free energy. The $\sum_{t,\mathbf{x}_{t-1},\mathbf{x}_{t},v_{t}}$
is an abbreviation for summing over $t$, $\mathbf{x}_{t-1}$, $\mathbf{x}_{t}$,
and $v_{t}$. $\sum_{\{\mathbf{x}_{t}\backslash x_{t}^{(m)}\}}$ is
the sum over all individuals in $\mathbf{x_{t}}$ except $x_{t}^{(m)}$.
We use similar abbreviations below. The first two sets of constraints
are marginalization conditions, and the third is normalization conditions.
To solve this constrained optimization problem, we first define the
Lagrange function using Lagrange multipliers to weight constraints,
then take the partial derivatives with respect to $\hat{\xi}(\mathbf{x}_{t-1},\mathbf{x}_{t},v_{t})$,
and $\hat{\gamma}^{(m)}(x_{t}^{(m)})$. The dual problem is to find
the approximate forward statistics $\hat{\alpha}_{t-1}^{(m)}(x_{t-1}^{(m)})$
and backward statistics $\hat{\beta}_{t}^{(m)}(x_{t}^{(m)})$ in order
to maximize the pseudo-likelihood function. The duality is between
minimizing Bethe free energy and maximizing pseudo-likelihood. The
fixed-point solution for the primal problem is as follows\footnote{The derivations for the optimization problem and its solution are
shown in the Supplemental Material.}: 
\begin{eqnarray}
\hat{\xi}(x_{t-1}^{(m)},x_{t}^{(m)},v_{t})=\frac{1}{Z_{t}}\sum_{m'\neq m,x_{t-1}^{(m')},x_{t}^{(m')}\hidewidth}{\scriptstyle P(\mathbf{x}_{t},v_{t}|\mathbf{x}_{t-1})\cdot\prod_{m}\hat{\alpha}_{t-1}^{(m)}(x_{t-1}^{(m)})\cdot\prod_{m}P(y_{t}^{(m)}|x_{t}^{(m)})\cdot\prod_{m}\hat{\beta}_{t}^{(m)}(x_{t}^{(m)})}.\label{eq:SKM2Slice}
\end{eqnarray}
$\hat{\xi}(x_{t-1}^{(m)},x_{t}^{(m)},v_{t})$ is the two-slice statistics
for an individual $m$, and $Z_{t}$ is the normalization constant.
Given the factorized form of $P(\mathbf{x}_{t},v_{t}|\mathbf{x}_{t-1})$
in Eqs.~(\ref{eq:SKMEventRate}) and (\ref{eq:SKMNERate}), everything
in Eq.~(\ref{eq:SKM2Slice}) can be written in a factorized form.
After reformulating the term relevant to the individual $m$, $\hat{\xi}(x_{t-1}^{(m)},x_{t}^{(m)},v_{t})$
can be shown neatly as follows: 
\begin{align}
 & \hat{\xi}_{t}(x_{t-1}^{(m)},x_{t}^{(m)},v_{t})=\frac{1}{Z_{t}}\hat{P}(x_{t}^{(m)},v_{t}|x_{t-1}^{(m)})\cdot\hat{\alpha}_{t-1}^{(m)}(x_{t-1}^{(m)})P(y_{t}^{(m)}|x_{t}^{(m)})\hat{\beta}_{t}^{(m)}(x_{t}^{(m)}),\label{eq:marginal2Slice}
\end{align}
where the marginalized transition kernel $\hat{P}(x_{t}^{(m)},v_{t}|x_{t-1}^{(m)})$
for the individual $m$ can be defined as:
\begin{align}
 & \hat{P}(x_{t}^{(m)},v_{t}=k|x_{t-1}^{(m)})={\displaystyle c_{k}g_{k}^{(m)}(x_{t-1}^{(m)})\prod\limits _{m'\ne m}\tilde{g}_{k,t-1}^{(m')}\cdot\delta(x_{t}^{(m)}-x_{t-1}^{(m)}\equiv\Delta_{k}^{(m)})},\label{eq:marginalEventRate}\\
 & \hat{P}(x_{t}^{(m)},v_{t}=\emptyset|x_{t-1}^{(m)})={\scriptstyle {\displaystyle \left(1-\sum\limits _{k}c_{k}g_{k}^{(m)}(x_{t-1}^{(m)})\prod\limits _{m'\ne m}\hat{g}_{k,t-1}^{(m')}\right)\delta(x_{t}^{(m)}-x_{t-1}^{(m)}\equiv0),}}\label{eq:marginalNoEventRate}\\
 & {\scriptstyle \tilde{g}_{k,t-1}^{(m')}=\sum\limits _{x_{t}^{(m')}-x_{t-1}^{(m')}\equiv\Delta_{k}^{(m')}\hidewidth}\alpha_{t-1}^{(m')}(x_{t-1}^{(m')})P(y_{t}^{(m')}|x_{t}^{(m')})\beta_{t}^{(m')}(x_{t}^{(m')})g_{k}^{(m')}(x_{t-1}^{(m')})\big/\sum\limits _{x_{t}^{(m')}-x_{t-1}^{(m')}\equiv0\hidewidth}\alpha_{t-1}^{(m')}(x_{t-1}^{(m')})P(y_{t}^{(m')}|x_{t}^{(m')})\beta_{t}^{(m')}(x_{t}^{(m')})},\nonumber \\
 & {\scriptstyle \hat{g}_{k,t-1}^{(m')}=\sum\limits _{x_{t}^{(m')}-x_{t-1}^{(m')}\equiv0\hidewidth}\alpha(x_{t-1}^{(m')})P(y_{t}^{(m')}|x_{t}^{(m')})\beta_{t}^{(m')}(x_{t}^{(m')})g_{k}^{(m')}(x_{t-1}^{(m')})\big/\sum\limits _{x_{t}^{(m')}-x_{t-1}^{(m')}\equiv0\hidewidth}\alpha_{t-1}^{(m')}(x_{t-1}^{(m')})P(y_{t}^{(m')}|x_{t}^{(m')})\beta_{t}^{(m')}(x_{t}^{(m')})},\nonumber 
\end{align}
In the above equations, we consider the mean field effect by summing
over the current and previous states of all the other individuals
$m'\ne m$. The marginalized transition kernel considers the probability
of event $k$ on the individual $m$ given the context of the temporal
evolutions of the other individuals. Comparing Eqs. (\ref{eq:marginalEventRate})
and (\ref{eq:marginalNoEventRate}) with Eqs. (\ref{eq:SKMEventRate})
and (\ref{eq:SKMNERate}), instead of multiplying $g_{k}^{(m')}(x_{t-1}^{(m')})$
for individual $m'\neq m$, we use the expected value of $g_{k}^{(m')}$
with respect to the marginal probability distribution of $x_{t-1}^{(m')}$.

\textbf{Complexity Analysis}: In our inference algorithm, the most
computation-intensive step is the marginalization in Eqs.~(\ref{eq:marginalEventRate})-(\ref{eq:marginalNoEventRate}).
The complexity is $O(MS^{2})$, where $M$ is the number of individuals
and $S$ is the state space of a single individual. The complexity
of the entire algorithm is therefore $O(MS^{2}TN)$, where $T$ is
the number of time steps and $N$ is the number of iterations until
convergence. As such, the complexity of our algorithm grows only linearly
with the number of individuals; it offers excellent scalability when
the number of tracked individuals becomes large. \vspace{-10pt}

\subsection{Parameter Learning\vspace{-5pt}
}

In order to learn the rate constant $c_{k}$, we maximize the expected
log likelihood. In a stochastic kinetic model, the probability of
a sample path is given in Eq. (\ref{eq:DTSKM}). The expected log
likelihood over the posterior probability conditioned on the observations
$\mathbf{y}_{1},\dots,\mathbf{y}_{T}$ takes the following form: 
\begin{align*}
\log P\left(\mathbf{x}_{1,\dots,T},\mathbf{y}_{1,\dots,T},v_{1,\dots,T}\right)=\sum_{t,\mathbf{x}_{t-1},\mathbf{x}_{t},v_{t}\hidewidth}\hat{\xi}_{t}(\mathbf{x}_{t-1},\mathbf{x}_{t},v_{t})\cdot\log(P(\mathbf{x}_{t},v_{t}|\mathbf{x}_{t-1})P(\mathbf{y}_{t}|\mathbf{x}_{t})).
\end{align*}
$\hat{\xi}_{t}\left(\mathbf{x}_{t-1},\mathbf{x}_{t},v_{t}\right)$
is the approximate two-slice statistics defined in Eq. (\ref{eq:BetheEnergy}).
Maximizing this expected log likelihood by setting its partial derivative
over the rate constants to 0 gives the maximum expected log likelihood
estimation of these rate constants. 
\begin{align}
c_{k}=\frac{\sum_{t,\mathbf{x}_{t-1},\mathbf{x}_{t}}\hat{\xi}_{t}(\mathbf{x}_{t-1},\mathbf{x}_{t},v_{t}=k)}{\sum_{t,\mathbf{x}_{t-1},\mathbf{x}_{t}}\hat{\xi}_{t}(\mathbf{x}_{t-1},\mathbf{x}_{t},v_{t}=\emptyset)g_{k}(\mathbf{x}_{t-1})}\approx\frac{\sum_{t}\ \sum_{\mathbf{x}_{t-1},\mathbf{x}_{t}}\hat{\xi}_{t}(\mathbf{x}_{t-1},\mathbf{x}_{t},v_{t}=k)}{\sum_{t}\ \prod_{m}\sum_{x_{t-1}^{(m)}}\hat{\gamma}_{t-1}^{(m)}(x_{t-1}^{(m)})g_{k}^{(m)}(x_{t-1}^{(m)})}.\label{eq:rateConstant}
\end{align}
As such, the rate constant for event $k$ is the expected number of
times that this event has occurred divided by the total expected number
of times this event could have occurred. 

To summarize, we provide the variational inference algorithm below.

\rule{1\columnwidth}{1pt}

\textbf{{Algorithm: Variational Inference with a Stochastic Kinetic
Model \hspace*{\fill}}}

Given the observations $y_{t}^{(m)}$ for $t=1,\dots,T$ and $m=1,\dots,M$,
find $x_{t}^{(m)}$, $v_{t}$ and rate constants $c_{k}$ for $k=1,\dots,V$.

\textbf{Latent state inference.} Iterate through the following forward
and backward passes until convergence, where $\hat{P}(x_{t}^{(m)},v_{t}|x_{t-1}^{(m)})$
is given by Eqs. (\ref{eq:marginalEventRate}) and (\ref{eq:marginalNoEventRate}). 
\begin{itemize}
\item Forward pass. For $t=1,\dots,T$ and $m=1,\dots,M$, update $\hat{\alpha}_{t}^{(m)}(x_{t}^{(m)})$
according to 
\begin{align*}
 & \hat{\alpha}_{t}^{(m)}(x_{t}^{(m)})\leftarrow\frac{1}{Z_{t}}\sum_{x_{t-1}^{(m)},v_{t}\hidewidth}\hat{\alpha}_{t-1}^{(m)}(x_{t-1}^{(m)})\hat{P}(x_{t}^{(m)},v_{t}|x_{t-1}^{(m)})P(y_{t}^{(m)}|x_{t}^{(m)}).
\end{align*}
\item Backward pass. For $t=T,\dots,1$ and $m=1,\dots,M$, update $\hat{\beta}_{t-1}^{(m)}(x_{t-1}^{(m)})$
according to 
\begin{align*}
 & \hat{\beta}_{t-1}^{(m)}(x_{t-1}^{(m)})\leftarrow\frac{1}{Z_{t}}\sum_{x_{t}^{(m)},v_{t}\hidewidth}\hat{\beta}_{t}^{(m)}(x_{t}^{(m)})\hat{P}(x_{t}^{(m)},v_{t}|x_{t-1}^{(m)})P(y_{t}^{(m)}|x_{t}^{(m)}).
\end{align*}
\end{itemize}
\textbf{Parameter estimation.} Iterate through the latent state inference
(above) and rate constants estimate of $c_{k}$ according to Eq. (\ref{eq:rateConstant}),
until convergence.

\rule{1\columnwidth}{1pt} \vspace{-20pt}

\section{Experiments on Epidemic Applications}

\label{sec:experiments} In this section, we evaluate the performance
of variational inference with a stochastic kinetic model (VISKM) algorithm
of epidemic dynamics, with which we predict the transmission of diseases
and the health status of each individual based on proximity data collected
from sensor networks. 

\subsection{Epidemic Dynamics}

In epidemic dynamics, $G_{t}=(\mathcal{M},E_{t})$ is a dynamic network,
where each node $m\in\mathcal{M}$ is an individual in the network,
and $E_{t}=\{(m_{i},m_{j})\}$ is a set of edges in $G_{t}$ representing
that individuals $m_{i}$ and $m_{j}$ have interacted at a specific
time $t$. There are two possible hidden states for each individual
$m$ at time $t$, $x_{t}^{(m)}\in\{0,1\}$, where 0 indicates the
susceptible state and 1 the infectious state. $y_{t}^{(m)}\in\{0,1\}$
represents the presence or absence of symptoms for individual $m$
at time $t$. $P(y_{t}^{(m)}|x_{t}^{(m)})$ represents the observation
probability. We define three types of events in epidemic applications:
(1) A previously infectious individual recovers and becomes susceptible
again: $I\overset{c_{1}}{\longrightarrow}S$. (2) An infectious individual
infects a susceptible individual in the network: $S+I\overset{c_{2}}{\longrightarrow}2I$.
(3) A susceptible individual in the network is infected by an outside
infectious individual: $S\overset{c_{3}}{\longrightarrow}I$. Based
on these events, the transition kernel can be defined as follows:
\begin{align*}
 & P(x_{t}^{(m)}=0|x_{t-1}^{(m)}=1)=c_{1},~P(x_{t}^{(m)}=1|x_{t-1}^{(m)}=1)=1-c_{1},\\
 & P(x_{t}^{(m)}\negthinspace=\negthinspace0|x_{t-1}^{(m)}\negthinspace=\negthinspace0)=(1-c_{3})(1-c_{2})^{C_{m,t}},~P(x_{t}^{(m)}\negthinspace=\negthinspace1|x_{t-1}^{(m)}\negthinspace=\negthinspace0)=1-(1-c_{3})(1-c_{2})^{C_{m,t}},
\end{align*}
where $C_{m,t}=\sum_{m':(m',m)\in E_{t}}\delta(x_{t}^{(m')}\equiv1)$
is the number of possible infectious sources for individual $m$ at
time $t$. Intuitively, the probability of a susceptible individual
becoming infected is 1 minus the probability that no infectious individuals
(inside or outside the network) infected him. When the probability
of infection is very small, we can approximate $P(x_{t}^{(m)}=1|x_{t-1}^{(m)}=0)\approx c_{3}+c_{2}\cdot C_{m,t}$. 

\subsection{Experimental Results}

\textbf{Data Explanation:} We employ two data sets of epidemic dynamics.
The real data set is collected from the Social Evolution experiment
\cite{DoLe11}. This study records ``common cold'' symptoms of 65
students living in a university residence hall from January 2009 to
April 2009, tracking their locations and proximities using mobile
phones. In addition, the students took periodic surveys regarding
their health status and personal interactions. The synthetic data
set was collected on the Dartmouth College campus from April 2001
to June 2004, and contains the movement history of 13,888 individuals
\cite{DaTr07}. We synthesized disease transmission along a timeline
using the popular susceptible-infectious-susceptible (SIS) epidemiology
model \cite{KeRo08}, then applied the VISKM to calibrate performance.
We selected this data set because we want to demonstrate that our
model works on data with a large number of people over a long period
of time.

\textbf{Evaluation Metrics and Baseline Algorithms:} We select the
receiver operating characteristic (ROC) curve as our performance metric
because the discrimination thresholds of diseases vary. 
We first compare the accuracy and efficiency of VISKM with Gibbs sampling
(Gibbs) and particle filtering (PF) on the Social Evolution data set
\cite{DoPe12,DoJo09}.\footnote{Code and data are available at \url{http://cse.buffalo.edu/~wendong/}.}
Both Gibbs sampling and particle filtering iteratively sample the
infectious and susceptible latent state sequences and the infection
and recovery events conditioned on these state sequences. Gibbs-Prediction-10000
indicates 10,000 iterations of Gibbs sampling with 1000 burn-in iterations
for the prediction task. PF-Smoothing-1000 similarly refers to 1000
iterations of particle filtering for the smoothing task. All experiments
are performed on the same computer.

\begin{figure*}
\centering \subfigure[Prediction]{\label{fig:mit_prediction_comp}\includegraphics[width=0.3\textwidth]{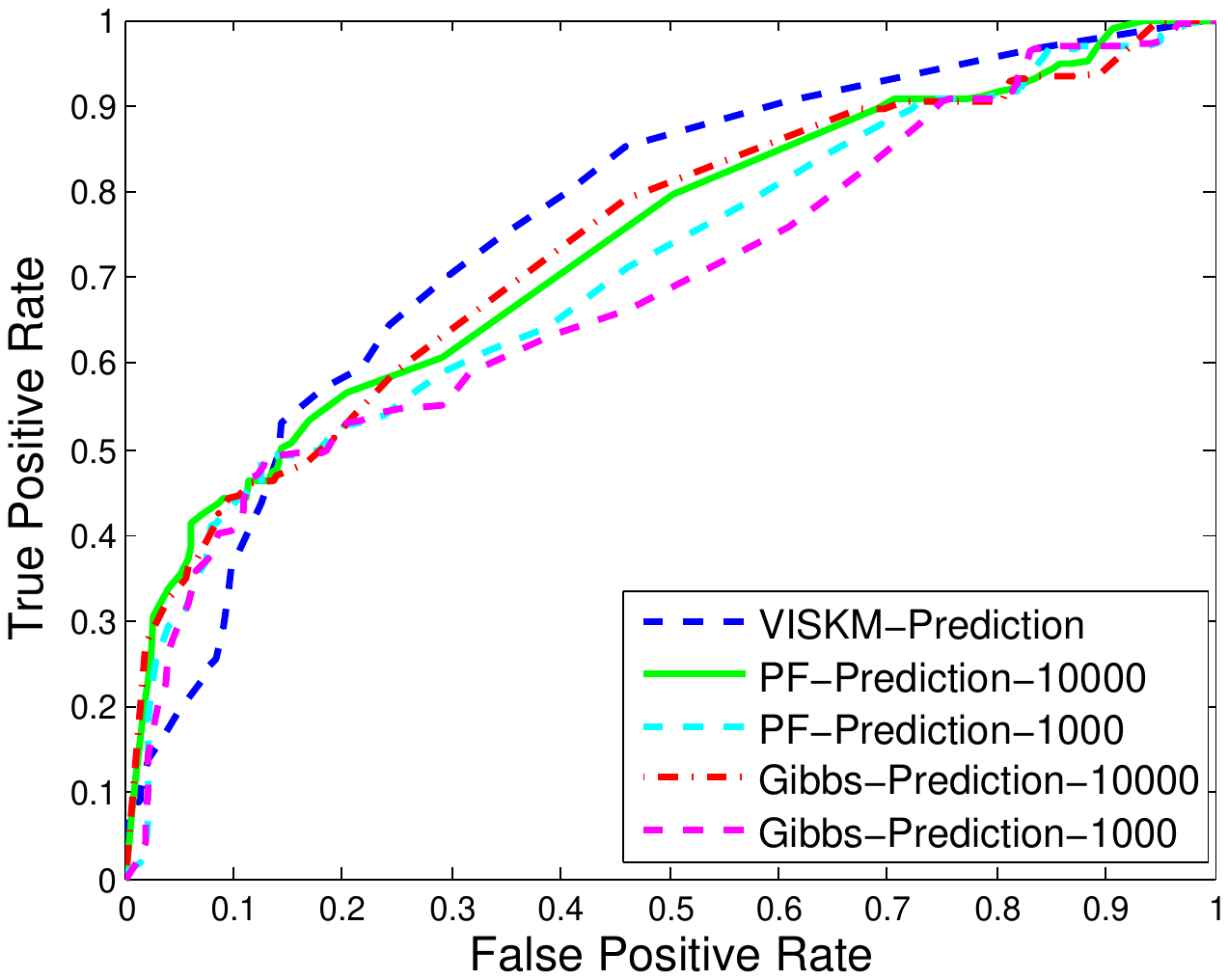}}
\subfigure[Smoothing]{\label{fig:mit_smoothing_comp}\includegraphics[width=0.3\textwidth]{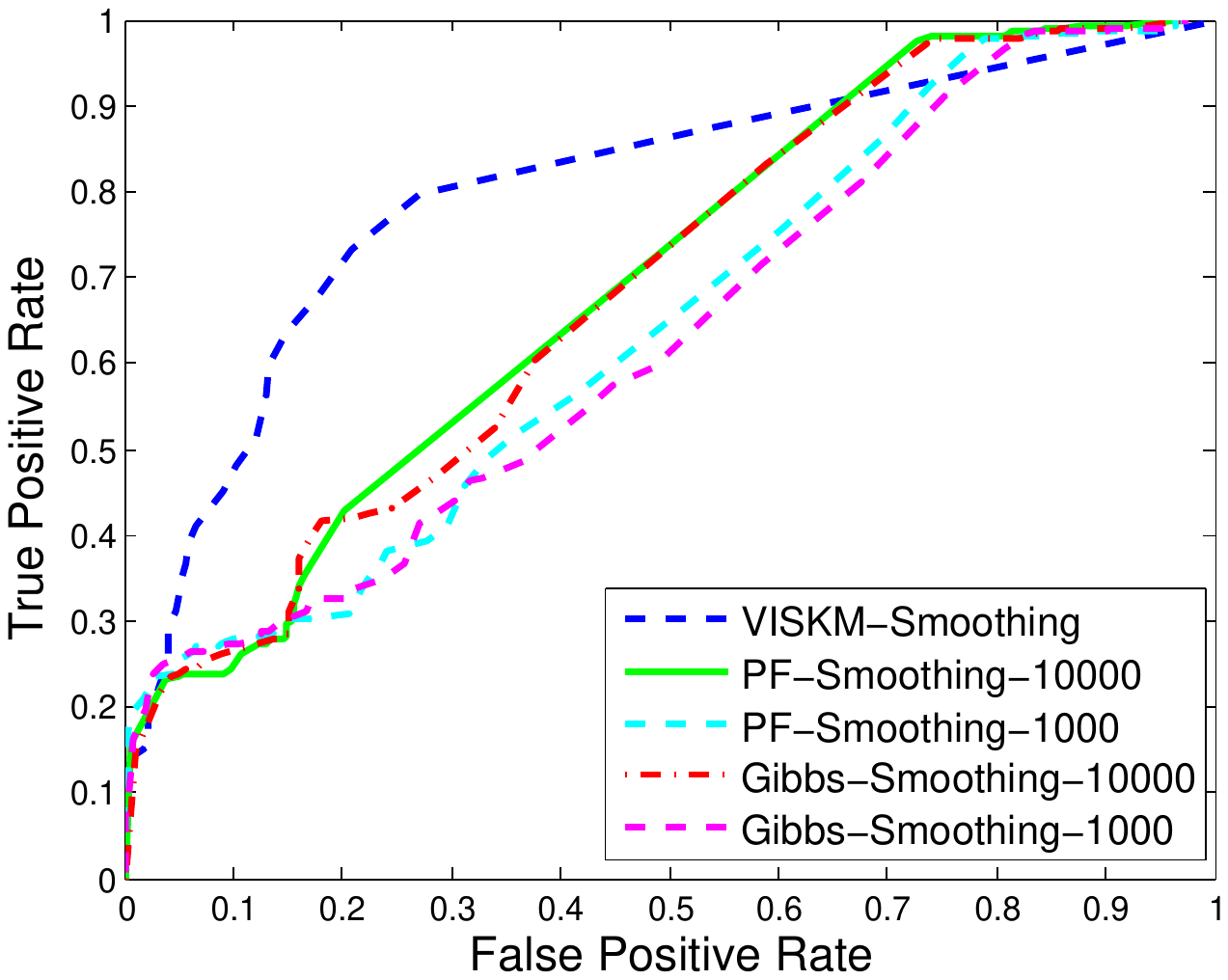}}
\subfigure[Expansion]{\label{fig:mit_expansion_comp}\includegraphics[width=0.3\textwidth]{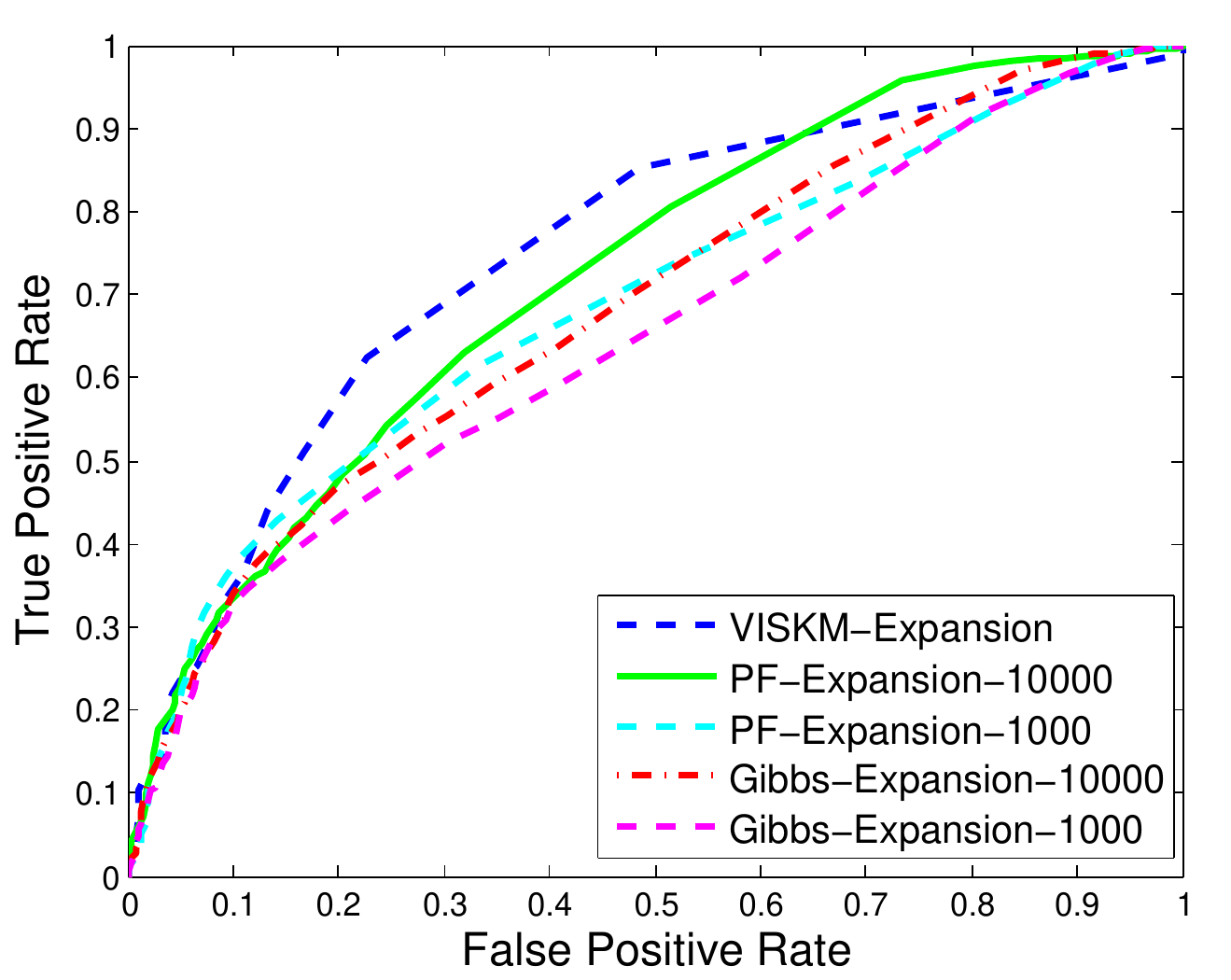}}
\subfigure[Dartmouth]{\label{fig:dart_three_tasks_comp}\includegraphics[width=0.3\textwidth]{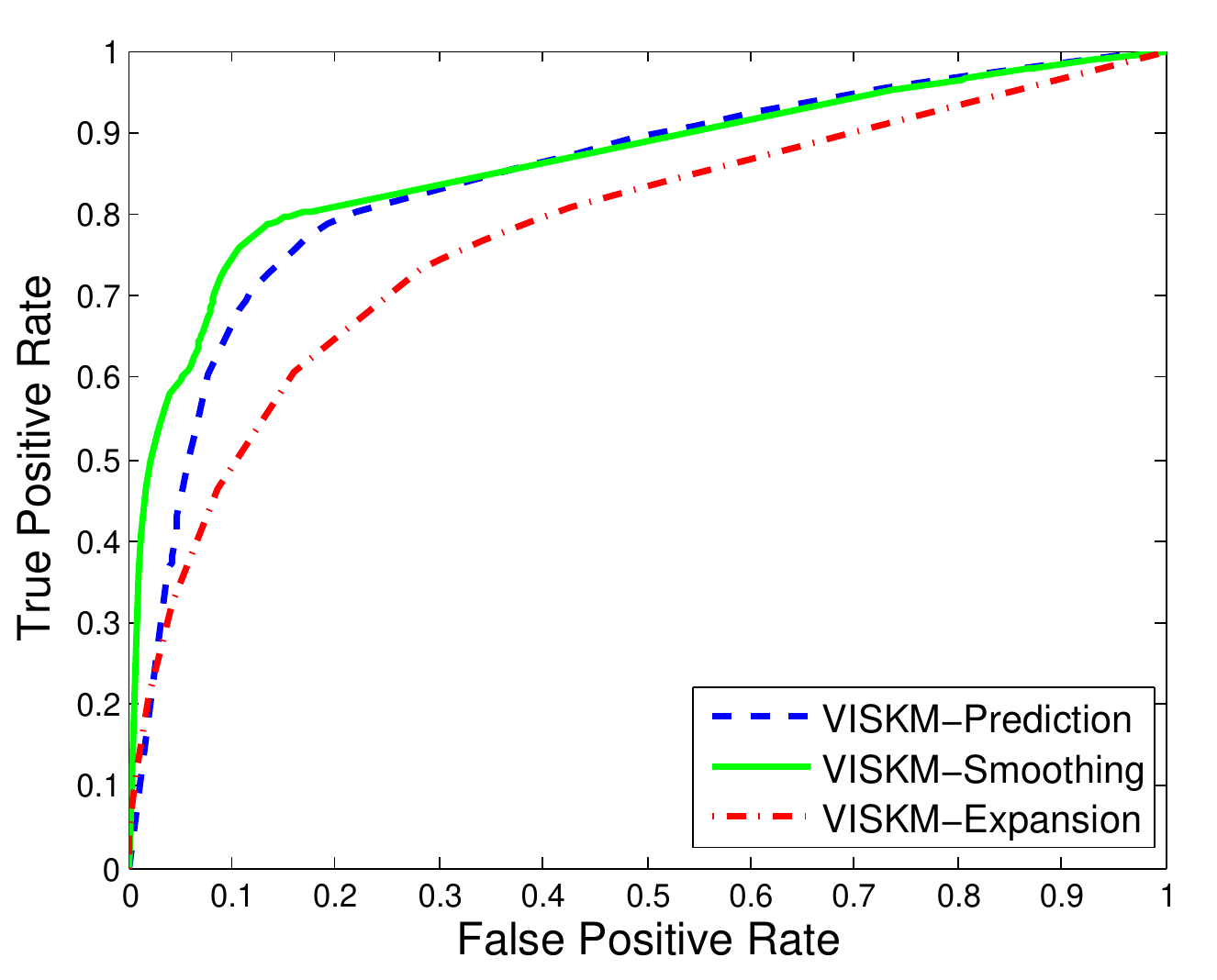}}
\subfigure[Social Evolution Statistics]{\label{fig:mit_stat}\includegraphics[width=0.3\textwidth]{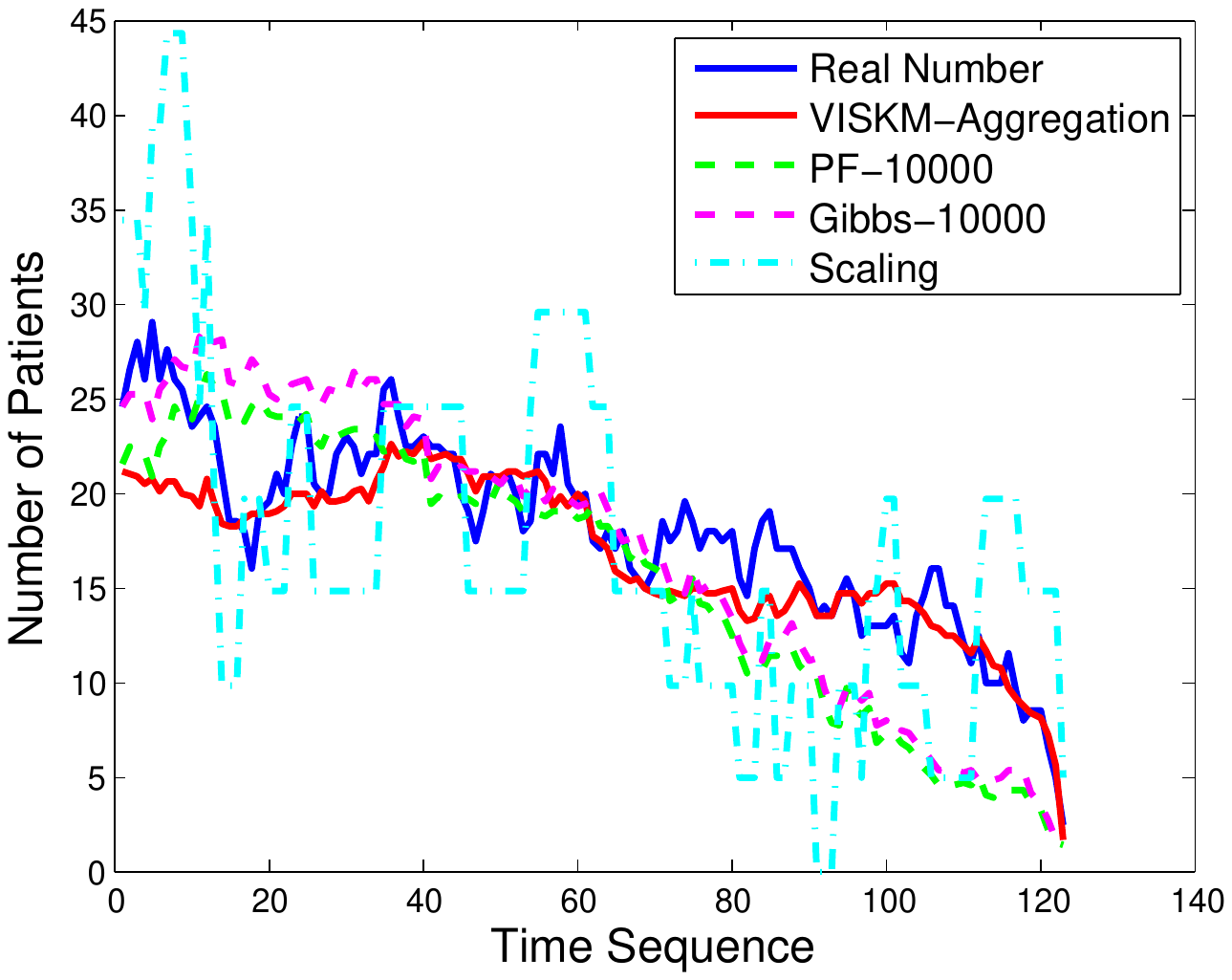}}
\subfigure[Dartmouth Statistics]{\label{fig:dart_stat}\includegraphics[width=0.3\textwidth]{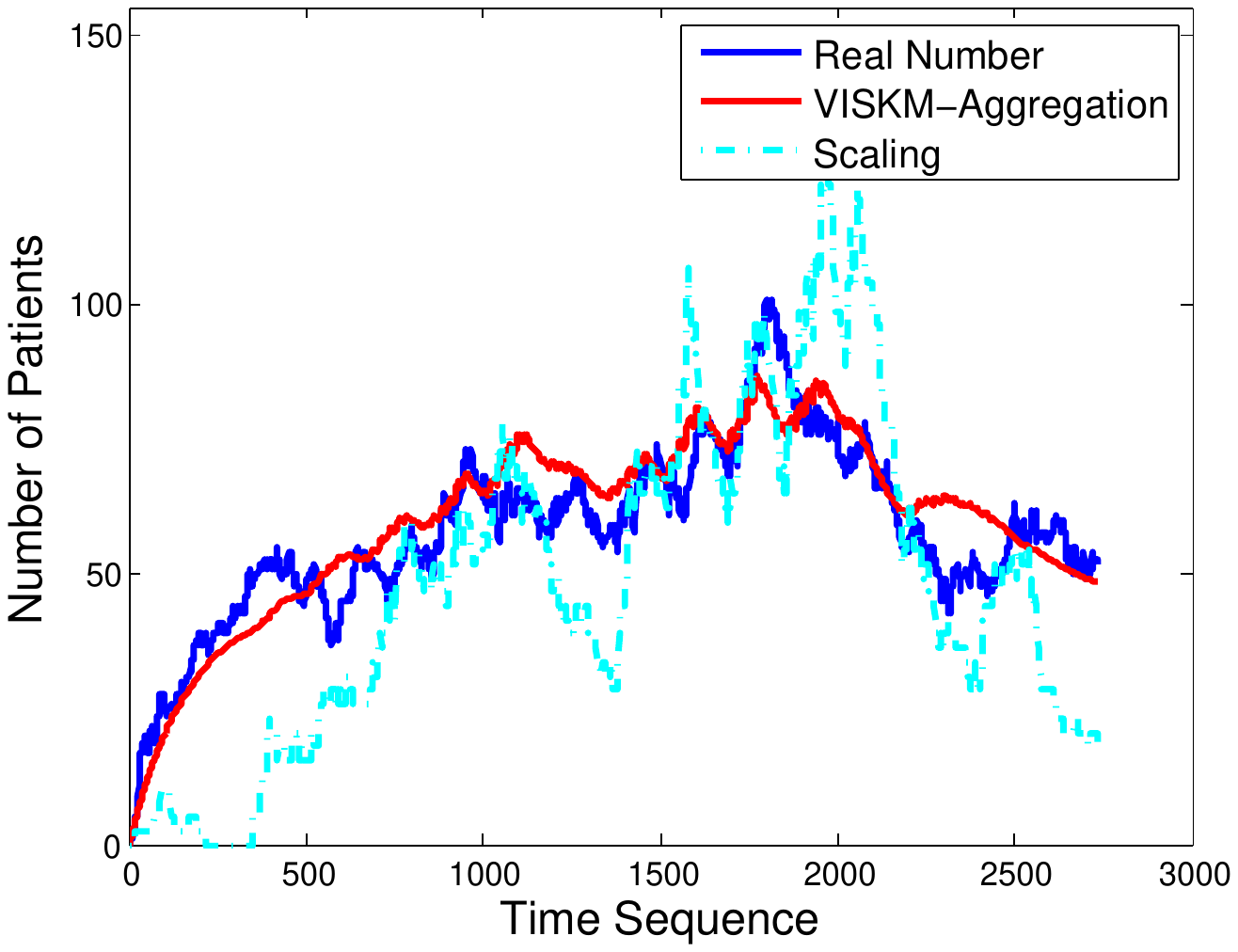}}
\protect\caption{Experimental results. (a-c) show the prediction, smoothing, and expansion
performance comparisons for Social Evolution data, while (d) shows
performance of the three tasks for Dartmouth data. (e-f) represent
the statistical inferences for both data sets.}
\label{fig:exp_result} 
\end{figure*}

\textbf{Individual State Inference:} We infer the probabilities of
a hidden infectious state for each individual at different times under
different scenarios. There are three tasks: 1. \emph{Prediction}:
Given an individual's past health and current interaction patterns,
we predict the current infectious latent state. Figure \ref{fig:exp_result}(a)
compares prediction performance among the different approximate inference
methods. 2. \emph{Smoothing:} Given an individual's interaction patterns
and past health with missing periods, we infer the infectious latent
states during these missing periods. Figure \ref{fig:exp_result}(b)
compares the performance of the three inference methods. 3. \emph{Expansion}:
Given the health records of a portion ($\sim10\%$) of the population,
we estimate the individual infectious states of the entire population
before medically inspecting them. For example, given either a group
of volunteers willing to report their symptoms or the symptom data
of patients who came to hospitals, we determine the probabilities
that the people near these individuals also became or will become
infected. This information helps the government or aid agencies to
efficiently distribute limited medical resources to those most in
need. Figure \ref{fig:exp_result}(c) compares the performance of
the different methods. From the above three graphs, we can see that
all three methods identify the infectious states in an accurate way.
However, VISKM outperforms Gibbs sampling and particle filtering in
terms of area under the ROC curve for all three tasks. VISKM has an
advantage in the smoothing task because the backward pass helps to
infer the missing states using subsequent observations. In addition,
the performance of Gibbs and PF improves as the number of samples/particles
increases. 

Figure \ref{fig:exp_result}(d) shows the performance of the three
tasks on the Dartmouth data set. We do not apply the same comparison
because it takes too much time for sampling. From the graph, we can
see that VISKM infers most of the infectious moments of individuals
in an accurate way for a large social system. In addition, the smoothing
results are slightly better than the prediction results because we
can leverage observations from both directions. The expansion case
is \emph{relatively} poor, because we use only very limited information
to derive the results; however, even in this case the ROC curve has
good discriminating power to differentiate between infectious and
susceptible individuals.

\textbf{Collective Statistics Inference:} After determining the individual
results, we aggregate them to approximate the total number of infected
individuals in the social system as time evolves. This offers a collective
statistical summary of the spread of disease in one area as in traditional
research, which typically scales the sample statistics with respect
to the sample ratio. Figures \ref{fig:exp_result}(e) and (f) show
that given $20\%$ of the Social Evolution data and $10\%$ of the
Dartmouth data, VISKM estimates the collective statistics better than
the other methods.

\textbf{Efficiency and Scalability:} Table \ref{Running_time} shows
the running time of different algorithms for the Social Evolution
data on the same computer. From the table, we can see that Gibbs sampling
runs slightly longer than PF, but they are in the same scale. However,
VISKM requires much less computation time. In addition, the computation
time of VISKM grows linearly with the number of individuals, which
validates the complexity analysis in Section \ref{sec:variational_inference}.
Thus, it offers excellent scalability for large social systems. In
comparison, Gibbs sampling and PF grow super linearly with the number
of individuals, and roughly linearly with the number of samples.

\textbf{Summary:} Our proposed VISKM achieves higher accuracy in terms
of area under ROC curve and collective statistics than Gibbs sampling
or particle filtering (within 10,000 iterations). More importantly,
VISKM is more efficient than sampling with much less computation time.
Additionally, the computation time of VISKM grows linearly with the
number of individuals, demonstrating its excellent scalability for
large social systems.

\begin{table}[!t]
\global\long\def\arraystretch{1.2}
 \centering \caption{Running time for different approximate inference algorithms. Gibbs\_10000
refers to Gibbs sampling for 10,000 iterations, and PF\_1000 to particle
filtering for 1000 iterations. Other entries follow the same pattern.
All times are measured in seconds.}
\label{Running_time} \begin{small} %
\begin{tabular}{|c|c|c|c|c|c|}
\hline 
~  & VISKM  & Gibbs\_1000  & Gibbs\_10000  & PF\_1000  & PF\_10000 \tabularnewline
\hline 
60 People  & 0.78  & 771  & 7820  & 601  & 6100 \tabularnewline
\hline 
30 People  & 0.39  & 255  & 2556  & 166  & 1888 \tabularnewline
\hline 
15 People  & 0.19  & 101  & 1003  & 122  & 1435 \tabularnewline
\hline 
\end{tabular}\end{small} 
\end{table}

\section{Conclusions}

In this paper, we leverage sensor network and social network data
to capture temporal evolution in social dynamics and infer individual
behaviors. In order to define the adaptive transition kernel, we introduce
a stochastic dynamic mode that captures the dynamics of complex interactions.
In addition, in order to make tractable inferences we propose a variational
inference algorithm the computation complexity of which grows linearly
with the number of individuals. Large-scale experiments on epidemic
dynamics demonstrate that our method effectively captures the evolution
of social dynamics and accurately infers individual behaviors. More
accurate collective effects can be also derived through the aggregated
results. Potential applications for our algorithm include the dynamics
of emotion, opinion, rumor, collaboration, and friendship.

\bibliographystyle{plain}

\newpage{}

\section{Appendix}

\subsection{Derivation of the optimization problem in Eq.(\ref{eq:BetheEnergy})}

Let $P(\mathbf{x}_{1,...,T},v_{1,...,T}|\mathbf{y}_{1,...,T})$ be
the exact posterior. Our goal is to approximate this posterior by
a distribution $Q(\mathbf{x}_{1,...,T},v_{1,...,T})$ in the exponential
family that minimizes the KL divergence between these two distributions:
\begin{align}
 & KL(Q(\mathbf{x}_{1,...,T},v_{1,...,T})|P(\mathbf{x}_{1,...,T},v_{1,...,T}|\mathbf{y}_{1,...,T}))\nonumber \\
= & \sum_{\mathbf{x}_{1,...,T},v_{1,...,T}}Q(\mathbf{x}_{1,...,T},v_{1,...,T})\log[\frac{Q(\mathbf{x}_{1,...,T},v_{1,...,T})\cdot P(\mathbf{y}_{1,...,T})}{P(\mathbf{x}_{1,...,T},\mathbf{y}_{1,...,T},v_{1,...,T})}]\nonumber \\
= & \sum_{\mathbf{x}_{1,...,T},v_{1,...,T}}Q(\mathbf{x}_{1,...,T},v_{1,...,T})\log Q(\mathbf{x}_{1,...,T},v_{1,...,T})\nonumber \\
 & -\sum_{t=1}^{T}\sum_{\mathbf{x}_{1,...,T},v_{1,...,T}}Q(\mathbf{x}_{1,...,T},v_{1,...,T})\log P(\mathbf{x}_{t},\mathbf{y}_{t},v_{t}|\mathbf{x}_{t-1}).\label{eq:obj}
\end{align}
In the first step, we apply the definition of conditional probability
and KL-divergence. In the second, we omit $P(\mathbf{y}_{1,...,T})$
because it is a constant in this optimization problem. In addition,
we decompose $P\left(\mathbf{x}_{1,\dots,T},\mathbf{y}_{1,\dots,T},v_{1,\dots,T}\right)=\prod_{t=1}^{T}P(\mathbf{x}_{t},\mathbf{y}_{t},v_{t}|\mathbf{x}_{t-1})$.

We then define the approximate two-slice statistics $\hat{\xi}(\mathbf{x}_{t-1},\mathbf{x}_{t},v_{t})$
and one-slice statistics $\hat{\gamma}(\mathbf{x}_{t})$. Both are
in the exponential family. In this context, we have $M$ individuals
in the system and the mean-field approximation can be shown as $\hat{\gamma}(\mathbf{x}_{t})=\prod_{m=1}^{N}\hat{\gamma}^{(m)}(x_{t}^{(m)})$,
where $\hat{\gamma}^{(m)}(x_{t}^{(m)})$ is the approximate one-slice
statistics for individual $m$. Given the observation that $Q(\mathbf{x}_{1,...,T},v_{1,...,T})$
can be expressed as a product of two-slice statistics divided by a
product of one-slice statistics, then 
\begin{align}
 & Q(\mathbf{x}_{1,...,T},v_{1,...,T})=\frac{\prod_{t=1}^{T}\hat{\xi}(\mathbf{x}_{t-1},\mathbf{x}_{t},v_{t})}{\prod_{t=1}^{T-1}\hat{\gamma}(\mathbf{x}_{t})}=\frac{\prod_{t=1}^{T}\hat{\xi}(\mathbf{x}_{t-1},\mathbf{x}_{t},v_{t})}{\prod_{t=1}^{T-1}\prod_{m=1}^{M}\hat{\gamma}^{(m)}(x_{t}^{(m)})}.\label{eq:decomp}
\end{align}
If we substitute Eq. (\ref{eq:decomp}) into Eq. (\ref{eq:obj}),
the objective function becomes the following: 
\begin{align}
 & \sum_{\mathbf{x}_{1,...,T},v_{1,...,T}}Q(\mathbf{x}_{1,...,T},v_{1,...,T})\log\frac{\prod_{t=1}^{T}\hat{\xi}(\mathbf{x}_{t-1},\mathbf{x}_{t},v_{t})}{\prod_{t=1}^{T-1}\prod_{m}\hat{\gamma}^{(m)}(x_{t}^{(m)})}\nonumber \\
 & -\sum_{t=1}^{T}\sum_{\mathbf{x}_{1,...,T},v_{1,...,T}}Q(\mathbf{x}_{1,...,T},v_{1,...,T})\log P(\mathbf{x}_{t},\mathbf{y}_{t},v_{t}|\mathbf{x}_{t-1})\nonumber \\
 & =\sum_{t,\mathbf{x}_{t-1},\mathbf{x}_{t},v_{t}}\hat{\xi}(\mathbf{x}_{t-1},\mathbf{x}_{t},v_{t})\log\frac{\hat{\xi}(\mathbf{x}_{t-1},\mathbf{x}_{t},v_{t})}{P(\mathbf{x}_{t},\mathbf{y}_{t},v_{t}|\mathbf{x}_{t-1})}\nonumber \\
 & -\sum_{t,\mathbf{x}_{t}}\prod_{m}\hat{\gamma}^{(m)}(x_{t}^{(m)})\log\prod_{m}\hat{\gamma}^{(m)}(x_{t}^{(m)}).\label{eq:mulfinal}
\end{align}
This objective function is subject to marginalization and normalization
constraints: 
\begin{align*}
 & \sum_{v_{t},\mathbf{x}_{t-1},\{\mathbf{x}_{t}\backslash x_{t}^{(m)}\}\hidewidth}\hat{\xi}(\mathbf{x}_{t-1},\mathbf{x}_{t},v_{t})=\hat{\gamma}_{t}^{(m)}(x_{t}^{(m)})\mbox{, for all }t,m,x_{t}^{(m)},\\
 & \sum_{v_{t},\{\mathbf{x}_{t-1}\backslash x_{t-1}^{(m)}\},\mathbf{x}_{t}\hidewidth}\hat{\xi}(\mathbf{x}_{t-1},\mathbf{x}_{t},v_{t})=\hat{\gamma}_{t-1}^{(m)}(x_{t-1}^{(m)})\mbox{, for all }t,m,x_{t-1}^{(m)},\\
 & \sum_{x_{t}^{(m)}\hidewidth}\hat{\gamma}_{t}^{(m)}(x_{t}^{(m)})=1\mbox{, for all }t,m.
\end{align*}
$\sum_{\{\mathbf{x}_{t}\backslash x_{t}^{(m)}\}}$ refers to the sum
over all values of $\mathbf{x}_{t}$ except $x_{t}^{(m)}$.

\newpage{}

\subsection{Derivation of the inference algorithm from Eq.(\ref{eq:marginal2Slice})
to Eq.(\ref{eq:marginalNoEventRate})}

The optimization problem derived from Eq. (\ref{eq:mulfinal}) along
with the constraints can be shown as follows: 
\begin{align}
 & \sum\limits _{t,\mathbf{x}_{t-1},\mathbf{x}_{t},v_{t}\hidewidth}\hat{\xi}(\mathbf{x}_{t-1},\mathbf{x}_{t},v_{t})\log\frac{\hat{\xi}(\mathbf{x}_{t-1},\mathbf{x}_{t},v_{t})}{P(\mathbf{x}_{t},\mathbf{y}_{t},v_{t}|\mathbf{x}_{t-1})}-\sum\limits _{t,\mathbf{x}_{t}}\prod\limits _{m}\hat{\gamma}_{t}^{(m)}(x_{t}^{(m)})\log\prod\limits _{m}\hat{\gamma}_{t}^{(m)}(x_{t}^{(m)})\label{eq:KL}\\
 & \mbox{subject to: }\nonumber \\
 & \sum_{v_{t},\mathbf{x}_{t-1},\{\mathbf{x}_{t}\backslash x_{t}^{(m)}\}\hidewidth}\hat{\xi}(\mathbf{x}_{t-1},\mathbf{x}_{t},v_{t})=\hat{\gamma}_{t}^{(m)}(x_{t}^{(m)})\mbox{, for all }t,m,x_{t}^{(m)},\nonumber \\
 & \sum_{v_{t},\{\mathbf{x}_{t-1}\backslash x_{t-1}^{(m)}\},\mathbf{x}_{t}\hidewidth}\hat{\xi}(\mathbf{x}_{t-1},\mathbf{x}_{t},v_{t})=\hat{\gamma}_{t-1}^{(m)}(x_{t-1}^{(m)})\mbox{, for all }t,m,x_{t-1}^{(m)},\nonumber \\
 & \sum_{x_{t}^{(m)}\hidewidth}\hat{\gamma}_{t}^{(m)}(x_{t}^{(m)})=1\mbox{, for all }t,m.\nonumber 
\end{align}

We apply the method of Lagrange multipliers to solve this, which begins
with forming the Lagrange function to be optimized:

\begin{align}
L & =\sum\limits _{t,\mathbf{x}_{t-1},\mathbf{x}_{t},v_{t}\hidewidth}\hat{\xi}(\mathbf{x}_{t-1},\mathbf{x}_{t},v_{t})\log\frac{\hat{\xi}(\mathbf{x}_{t-1},\mathbf{x}_{t},v_{t})}{P(\mathbf{x}_{t},\mathbf{y}_{t},v_{t}|\mathbf{x}_{t-1})}-\sum\limits _{t,\mathbf{x}_{\mathbf{t}}}\prod\limits _{m}\hat{\gamma}_{t}^{(m)}(x_{t}^{(m)})\log\prod\limits _{m}\hat{\gamma}_{t}^{(m)}(x_{t}^{(m)})\label{eq:Lagrange}\\
 & +\sum_{\hidewidth t,m,x_{t}^{(m)}}\lambda_{t}^{(m)}(x_{t}^{(m)})\left(\sum_{v_{t},\mathbf{x}_{t-1},\{\mathbf{x}_{t}\backslash x_{t}^{(m)}\}\hidewidth}\hat{\gamma}_{t}^{(m)}(x_{t}^{(m)})-\hat{\xi}(\mathbf{x}_{t-1},\mathbf{x}_{t},v_{t})\right)\nonumber \\
 & +\sum_{\hidewidth t,m,x_{t-1}^{(m)}}\mu_{t-1}^{(m)}(x_{t-1}^{(m)})\left(\sum_{v_{t},\{\mathbf{x}_{t-1}\backslash x_{t-1}^{(m)}\},\mathbf{x}_{t}\hidewidth}\hat{\gamma}_{t-1}^{(m)}(x_{t-1}^{(m)})-\hat{\xi}(\mathbf{x}_{t-1},\mathbf{x}_{t},v_{t})\right).\nonumber \\
 & +\sum_{\hidewidth t,m,x_{t}^{(m)}}\nu(x_{t}^{(m)})\left(\sum_{x_{t}^{(m)}\hidewidth}\hat{\gamma}_{t}^{(m)}(x_{t}^{(m)})-1\right)\nonumber 
\end{align}
We then set the partial derivatives of Eq.~(\ref{eq:Lagrange}) over
$\hat{\xi}(\mathbf{x}_{t-1},\mathbf{x}_{t},v_{t})$ to 0, which results
in the following: 
\begin{align*}
 & \frac{\partial L}{\partial\hat{\xi}(\mathbf{x}_{t-1},\mathbf{x}_{t},v_{t})}=\log\frac{\hat{\xi}(\mathbf{x}_{t-1},\mathbf{x}_{t},v_{t})}{P(\mathbf{x}_{t},\mathbf{y}_{t},v_{t}|\mathbf{x}_{t-1})}+1-\sum_{m}\lambda_{t}^{(m)}(x_{t}^{(m)})-\sum\limits _{m}\mu_{t-1}^{(m)}(x_{t-1}^{(m)})\stackrel{\mbox{set}}{=}0\\
 & \Rightarrow\hat{\xi}(\mathbf{x}_{t-1},\mathbf{x}_{t},v_{t})\propto\exp\left(\sum\limits _{m}\mu_{t-1}^{(m)}(x_{t-1}^{(m)})\right)P(\mathbf{x}_{t},\mathbf{y}_{t},v_{t}|\mathbf{x}_{t-1})\exp\left(\sum\limits _{m}\lambda_{t}^{(m)}(x_{t}^{(m)})\right),\\
\end{align*}
As such, we see that ${\hat{\alpha}_{t-1}^{(m)}(x_{t-1}^{(m)})=\exp(\mu_{t-1}^{(m)}(x_{t-1}^{(m)}))}$
is associated with the forward probabilities and ${\hat{\beta}_{t}^{(m)}(x_{t}^{(m)})=\exp(\lambda_{t}^{(m)}(x_{t}^{(m)}))}$
with the backward probabilities, with $\hat{\gamma}_{t}^{(m)}(x_{t}^{(m)})=\hat{\alpha}_{t}^{(m)}(x_{t}^{(m)})\hat{\beta}_{t}^{(m)}(x_{t}^{(m)})$.
We can determine the two-slice statistics for an individual $m$ by
marginalizing the other individuals $m'\neq m$: 
\begin{align*}
 & \hat{\xi}(x_{t-1}^{(m)},x_{t}^{(m)},v_{t})=\sum_{m'\neq m,x_{t-1}^{(m')},x_{t}^{(m')}\hidewidth}\hat{\xi}(\mathbf{x}_{t-1},\mathbf{x}_{t},v_{t})\\
 & \propto\sum_{m'\neq m,x_{t-1}^{(m')},x_{t}^{(m')}\hidewidth}P(\mathbf{x}_{t},v_{t}|\mathbf{x}_{t-1})\cdot\prod_{m}\hat{\alpha}_{t-1}^{(m)}(x_{t-1}^{(m)})\cdot\prod_{m}P(y_{t}^{(m)}|x_{t}^{(m)})\cdot\prod_{m}\hat{\beta}_{t}^{(m)}(x_{t}^{(m)}).
\end{align*}
The above is the same as in Eq.~(\ref{eq:SKM2Slice}).

\subsection{Derivation of the parameter-learning algorithm}

From Eq.(\ref{eq:DTSKM}), the log-likelihood of the entire sequence
can be shown as this: 
\begin{align}
 & \log P\left(\mathbf{x}_{1,\dots,T},\mathbf{y}_{1,\dots,T},v_{1,\dots,T}\right)=\sum_{t=1}^{T}\log P(\mathbf{x}_{t},v_{t}|\mathbf{x}_{t-1})+\sum_{t=1}^{T}\log P(\mathbf{y}_{t}|\mathbf{x}_{t}),\mbox{ where }\label{eq:log-likelihood}\\
 & P(\mathbf{x}_{t},v_{t}|\mathbf{x}_{t-1})=\begin{cases}
c_{k}\cdot g_{k}\left(\mathbf{x}_{t-1}\right)\cdot\delta(\mathbf{x}_{t}-\mathbf{x}_{t-1}\equiv\mathbf{\Delta_{k}}) & \mbox{if }v_{t}=k\\
(1-\sum_{k}c_{k}g_{k}\left(\mathbf{x}_{t-1}\right))\cdot\delta(\mathbf{x}_{t}-\mathbf{x}_{t-1}\equiv\mathbf{0}) & \mbox{if }v_{t}=\emptyset
\end{cases}.\nonumber 
\end{align}
The probabilities for state transition can be shown as the probabilities
of a set of events. The expected log likelihood over the posterior
probability conditioned on the observations $\mathbf{y}_{1},\dots,\mathbf{y}_{T}$
takes the following form: 
\begin{align}
 & \mathbf{E}_{P(\mathbf{x}_{1,...,T},v_{1,...,T}|\mathbf{y}_{1,...,T})}\left(\log P\left(\mathbf{x}_{1,\dots,T},\mathbf{y}_{1,\dots,T},v_{1,\dots,T}\right)\right)~~~~~~~~~~~~~~~~~~~~~~~~~~~~~~~~~~~~~~~~~~~~~\label{eq:expected-likelihood}\\
= & \sum_{t,\mathbf{x}_{t-1},\mathbf{x}_{t},v_{t}\hidewidth}\hat{\xi}_{t}(\mathbf{x}_{t-1},\mathbf{x}_{t},v_{t})\cdot\log\left(P(\mathbf{x}_{t},v_{t}|\mathbf{x}_{t-1})P(\mathbf{y}_{t}|\mathbf{x}_{t})\right)\nonumber \\
= & \sum_{t,\mathbf{x}_{t-1},\mathbf{x}_{t}\hidewidth}\hat{\xi}_{t}(\mathbf{x}_{t-1},\mathbf{x}_{t},v_{t}=v)\cdot\log\left(P(\mathbf{x}_{t},v_{t}=v|\mathbf{x}_{t-1})P(\mathbf{y}_{t}|\mathbf{x}_{t})\right)\nonumber \\
+ & \sum_{t,\mathbf{x}_{t-1},\mathbf{x}_{t}\hidewidth}\hat{\xi}_{t}(\mathbf{x}_{t-1},\mathbf{x}_{t},v_{t}=\emptyset)\cdot\log\left(P(\mathbf{x}_{t},v_{t}=\emptyset|\mathbf{x}_{t-1})P(\mathbf{y}_{t}|\mathbf{x}_{t})\right)\nonumber 
\end{align}
At a given time $t$, there are two possible cases: $v_{t}=v$, where
$v\in\{1,\cdots,V\}$, and $v_{t}=\emptyset$. The derivatives with
respect to $c_{k}$ can be shown as follows: 
\begin{align*}
 & \frac{\partial\log P(\mathbf{x}_{t},v_{t}=k|\mathbf{x}_{t-1})}{\partial c_{k}}=\frac{1}{c_{k}}\\
 & \frac{\partial\log P(\mathbf{x}_{t},v_{t}=\emptyset|\mathbf{x}_{t-1})}{\partial c_{k}}=\frac{-g_{k}(\mathbf{x}_{t-1})}{1-\sum_{k}c_{k}g_{k}(\mathbf{x}_{t-1})}
\end{align*}
Note that here we do not detail $\delta(\mathbf{x}_{t}-\mathbf{x}_{t-1}\equiv\mathbf{\Delta_{k}})$
and $\delta(\mathbf{x}_{t}-\mathbf{x}_{t-1}\equiv\mathbf{0})$ explicitly,
because when calculating the derivatives of expected log likelihood
in Eq.(\ref{eq:expected-likelihood}) these terms will be contained
in $\hat{\xi}_{t}(\mathbf{x}_{t-1},\mathbf{x}_{t},v_{t}=k)$ and $\hat{\xi}_{t}(\mathbf{x}_{t-1},\mathbf{x}_{t},v_{t}=\emptyset)$.
Next we take the derivative of expected log likelihood with respect
to $c_{k}$: 
\begin{align}
 & \frac{\mathbf{E}_{P(\mathbf{x}_{1,...,T},v_{1,...,T}|\mathbf{y}_{1,...,T})}\left(\log P\left(\mathbf{x}_{1,\dots,T},\mathbf{y}_{1,\dots,T},v_{1,\dots,T}\right)\right)}{\partial c_{k}}\\
= & \sum_{t,\mathbf{x}_{t-1},\mathbf{x}_{t}\hidewidth}\hat{\xi}_{t}(\mathbf{x}_{t-1},\mathbf{x}_{t},v_{t}=k)\frac{1}{c_{k}}-\sum_{t,\mathbf{x}_{t-1},\mathbf{x}_{t},\hidewidth}\hat{\xi}_{t}(\mathbf{x}_{t-1},\mathbf{x}_{t},v_{t}=\emptyset)\frac{g_{k}(\mathbf{x}_{t-1})}{1-\sum_{k}c_{k}g_{k}(\mathbf{x}_{t-1})}\nonumber 
\end{align}
Because we assume that the auxiliary event dominates when the time
step is small, we approximate $1-\sum_{k}c_{k}g_{k}(\mathbf{x}_{t})\approx1$
and $\sum_{\mathbf{x}_{t}}\hat{\xi}_{t}(\mathbf{x}_{t-1},\mathbf{x}_{t},v_{t}=\emptyset)\approx\hat{\gamma}_{t-1}(\mathbf{x}_{t-1})$.
After applying this approximation and setting the derivative to $0$,
the result is as follows: 
\begin{align}
c_{k} & =\frac{\sum_{t}\ \sum_{\mathbf{x}_{t-1},\mathbf{x}_{t}}\hat{\xi}_{t}(\mathbf{x}_{t-1},\mathbf{x}_{t},v_{t}=k)}{\sum_{t}\ \sum_{\mathbf{x}_{t-1},\mathbf{x}_{t}}\hat{\xi}_{t}(\mathbf{x}_{t-1},\mathbf{x}_{t},v_{t}=\emptyset)g_{k}(\mathbf{x}_{t-1})}\\
 & \approx\frac{\sum_{t}\ \sum_{\mathbf{x}_{t-1},\mathbf{x}_{t}}\hat{\xi}_{t}(\mathbf{x}_{t-1},\mathbf{x}_{t},v_{t}=k)}{\sum_{t}\ \sum_{\mathbf{x}_{t-1}}\hat{\gamma}_{t-1}(\mathbf{x}_{t-1})g_{k}(\mathbf{x}_{t-1})}\nonumber \\
 & =\frac{\sum_{t}\ \sum_{\mathbf{x}_{t-1},\mathbf{x}_{t}}\hat{\xi}_{t}(\mathbf{x}_{t-1},\mathbf{x}_{t},v_{t}=k)}{\sum_{t}\ \prod_{m}\sum_{x_{t-1}^{(m)}}\hat{\gamma}_{t-1}^{(m)}(x_{t-1}^{(m)})g_{k}^{(m)}(x_{t-1}^{(m)})}.\nonumber 
\end{align}

\end{document}